\pdfoutput=1

\documentclass[11pt]{article}

\usepackage[]{ACL2023}

\usepackage{enumitem}
\usepackage{color} 
\usepackage{graphicx}
\usepackage{stfloats}
\usepackage{booktabs}
\usepackage{subfigure}
\usepackage{todonotes}
\usepackage{array}
\usepackage{amssymb, amsmath}
\usepackage{multicol}
\usepackage{multirow}
\usepackage{threeparttable}
\usepackage{algorithm}
\usepackage{algorithmic}
\usepackage{microtype}
\usepackage{makecell}

\usepackage[T1]{fontenc}

\usepackage[utf8]{inputenc}

\usepackage{microtype}

\usepackage{inconsolata}

\title{Dialog Action-Aware Transformer for Dialog Policy Learning}

\author{
  Huimin Wang$^{1*}$, Wai-Chung Kwan$^{2,3}$\thanks{\ \ Equal Contribution}, Kam-Fai Wong$^{2,3}$ \\
  $^1$Jarvis Lab, Tencent, Shenzhen, China \\
  $^2$The Chinese University of Hong Kong, Hong Kong, China \\
  $^3$MoE Key Laboratory of High Confidence Software Technologies, China \\
    \texttt{\{hmmmwang\}@tencent.com} \\
    \texttt{\{wckwan,kfwong\}@se.cuhk.edu.hk}
 }

\begin{document}
\maketitle
\begin{abstract}
Recent works usually address Dialog policy learning DPL by training a reinforcement learning (RL) agent to determine the best dialog action.  However, existing works on deep RL require a large volume of agent-user interactions to achieve acceptable performance. In this paper, we propose to make full use of the plain text knowledge from the pre-trained language model to accelerate the RL agent's learning speed. Specifically, we design a dialog action-aware transformer encoder (DaTrans), which integrates a new fine-tuning procedure named masked last action task to encourage DaTrans to be dialog-aware and distils action-specific features. Then, DaTrans is further optimized in an RL setting with ongoing interactions and evolves through exploration in the dialog action space toward maximizing long-term accumulated rewards. The effectiveness and efficiency of the proposed model are demonstrated with both simulator evaluation and human evaluation. 
\end{abstract}

\section{Introduction}
A task-oriented dialog system that can serve users on certain tasks has increasingly attracted research efforts. 
Dialog policy learning (DPL) aiming to determine the next abstracted system output plays a key role in pipeline task-oriented dialog systems \cite{kwan_survey_2023}. Recently, it has shown great potential for using reinforcement learning (RL) based methods to formulate DPL \cite{young2013pomdp, su2016continuously, peng2017composite}. A lot of progress is being made in demonstration-based efficient learning methods \cite{brys2015reinforcement,cederborg2015policy,wang2020learning,li2020rethinking,jhunjhunwala2020multi,geishauser2022dynamic}. Among these methods, dialog state tracking (DST), comprising all information required to determine the response, is an indispensable module. However, DST inevitably accumulates errors from each module of the system. 

Recent pre-trained language models (PLMs) gathering knowledge from the massive plain text show great potential for formulating DPL without DST. Recently, the studies on PLMs for dialog, including BERT-based dialog state tracking \cite{gulyaev2020goal} and GPT-2 based dialog generation \cite{peng2020few, yang2020ubar} are not centred on DPL. To this end, we proposed the \textbf{D}ialog \textbf{A}ction-oriented transformer encoder termed as \textbf{DaTrans}, for efficient dialog policy training. DaTrans is achieved by a dialog act-aware fine-tuning task, which encourages the model to distil the dialog policy logic. Specifically, rather than commonly used tasks, like predicting randomly masked words in the input (MLM task) and classifying whether the sentences are continuous or not (NSP task) \cite{devlin2018bert}, DaTrans is fine-tuned by predicting the masked last acts in the input action sequences (termed as MLA task). After that, DaTrans works as an RL agent which evolves toward maximizing long-term accumulated rewards through interacting with a user simulator. Following the traditional RL-based dialog policy learning framework, the main novelty of DaTrans is that it integrates a proposed dialog action-aware fine-tuning task (MLA), which helps to extract action-specific features from historical dialog action sequences to improve dialog policy learning. The empirical results prove the excellent performance of DaTrans. Our main contributions include 1) We propose the DaTrans that integrates the dialog act-aware fine-tuning task to extract the dialog policy logic from the plain text; 2) We validate the efficiency and effectiveness of the proposed model on a multi-domain benchmark with both simulator and human evaluation.



\begin{figure}[t]
\setlength{\belowcaptionskip}{-0.4cm}   
\centering
\includegraphics[width=1.0\columnwidth]{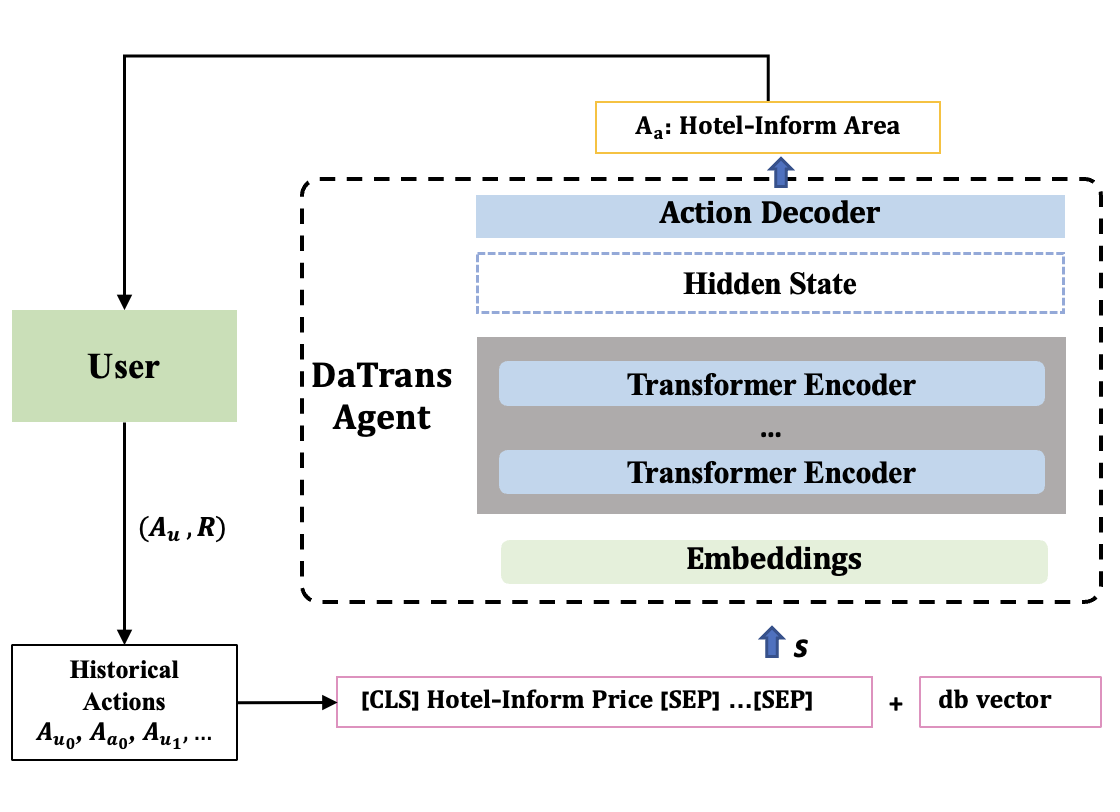}
\caption{The Illustration of \textbf{D}ialog \textbf{A}ction-oriented Transformer Encoder (\textbf{DaTrans}). In this example, \textbf{DaTrans} generates the dialog action $A_a$ based on historical actions.}
\label{fig:flow}
\end{figure}

\section{Approach}
We cast the dialog policy learning problem as a Markov Decision Process and optimize the policy with deep reinforcement learning approaches. RL usually involves an interactive process (as shown in Figure \ref{fig:flow}), during which the dialog agent’s behavior should choose actions that tend to increase the long-turn sum of rewards given by the user. It can learn to do this over time, by systematic trials and errors until reaches the optimal. In our setting, the dialog agent is encoded with the proposed DaTrans, which perceives the state $s$ and determines the next action $A_a$. We consider a transformer decoder-based policy model, which takes text concatenating of tuples containing a domain name, an intent type, and slot names as input and determines the next action.

\subsection{DaTrans}
We apply Deep Q-learning to optimize dialog policy. $Q_{\theta}(s,a)$, approximating the state-action value function parameterized $\theta$, is implemented based on DaTrans as illustrated in Figure \ref{fig:flow}. In each turn, perceiving the state $s$ that consists of historical action sequences and a database vector denoting the matches of the current constraints, DaTrans determines the dialog action $a$ with the generated value function $Q_\theta(\cdot|s)$. Historical action sequences are tokenized started from $\left[CLS\right]$, followed by the tokenized actions separated and ended with $\left[SEP\right]$. Then the transformer encoder gets the final hidden states denoted $\left[t_0.. t_n\right] = \textit{encoder}(\left[e_0..e_n\right])$ ($n$ is the current sequence length, $e_i$ is the embedding of the input token). The contextualized sentence-level representation $t_0$, is passed to a linear layer named action decoder $\boldsymbol{T}$ to generate: 
\begin{equation}
\begin{aligned}
   Q_{\theta}(s,a)=& \boldsymbol{T}_a (\textit{encoder}(\textit{Embed}(s)))
   \label{e:1}
\end{aligned}
\end{equation}
where $Embed$ is the embedding modules of transformer encoder, $\boldsymbol{T}_a$ denoted the $a_{th}$ output unit of $\boldsymbol{T}$. Based on DaTrans, the dialog policy is trained with $\epsilon$-greedy exploration that selects a random action with probability $\epsilon$, or adopts a greedy policy $a = argmax_{a'}Q_\theta(s, a')$. In each iteration, $Q_{\theta}(s,a)$ is updated by minimizing the following square loss with stochastic gradient descent:

\begin{equation}
\begin{aligned}
    \mathcal{L}_{\theta}&= \mathbb{E}_{(s,a, r, s') \sim D} [(y_i - Q_{\theta}(s,a))^2]\\
    y_i &= r + \gamma \max_{a'} Q'_{\theta}(s',a')
    \label{e:q-loss}
\end{aligned}
\end{equation}
where $\gamma \in \left[0, 1\right]$ is a discount factor, $D$ is the experience replay buffer with collected transition tuples $(s, a, r, s')$, $s$ is the current state,  $r$ refers to the reward, and $Q'(\cdot)$ is the target value function, which is only periodically updated, and $s^\prime$ is the next state. By differentiating the loss function with regard to $\theta$, we derive the following gradient:
\begin{equation}
\begin{aligned}
    \nabla_{\theta}\mathcal{L}(\theta) = 
    \mathbb{E}_{(s, a, r, s') 
    \sim D} 
    [( r +  \\ \gamma max_{a'}Q_{\theta'}^{'}(s',a') -  Q_{\theta}(s,a)) \nabla_{\theta}Q_{\theta}(s,a)]
\end{aligned}
\end{equation}
In each iteration, we update $Q(.)$ using minibatch Deep Q-learning.

\subsection{Dialog Action-aware Fine-tuning}
\label{sec:dialog Policy Oriented Fine-tuning}

A vanilla transformer decoder without pre-training can encumber the learning of dialog policy since it is totally unaware of the text and dialog logic. Meanwhile, well-pre-trained models like BERT, due to the generality of pre-training tasks and corpus, are still difficult with competent in dialog modeling. The NSP task encourages BERT to model the relationship between sentences, which may benefit natural language inference, however, biased dialog policy learning due to the inconsistency between success and continuity of sentences, e.g. discontinuous sentences can form a successful dialog. Also, the MLM task allows the word representation to fuse the left and right context, while the dialog agent is only allowed to access the left one. Considering that the ability to reason the next dialog action plays a key role in dialog policy, we replace the MLM and NSP task with a novel fine-tuning task: predicting masked last dialog action (MLA). MLA is based on a dialog action-aware fine-tuning corpus, each piece of which is a dialog session composed of the annotated historical action sequences, for example, “\textit{[CLS] Police-Inform Name [SEP] Police-Inform Phone Addr Post [SEP] general-thank none [SEP]}”, (denoted as \textbf{sentence A}). Then we randomly cut between two consecutive actions of a session, and select the first half with the masked last act as input. For example, we cut \textbf{sentence A} between the $2_{nd}$ and the $3_{rd}$ action, and mask the last act to get the input: "\textit{[CLS] Police-Inform Name [SEP] [MASK]..[MASK]}". The label for the masked tokens is "\textit{Police \ - \ Inform Phone Addr Post}". Significantly, the proposed MLA task for BERT is actually different from auto-regression. The way auto-regression works is after each token is produced, that token is added to the sequence of inputs and this new sequence becomes the input to the model in its next step. However, in DaTrans, the MLA task works as predicting the last dialog action
word by word without adding a new predicted word.

The goal of MLA is to minimize the cross-entropy loss with input tokens $w_0,w_1,..,w_n$:
\begin{equation}
\begin{aligned}
    \mathcal{L}^{mla} =& -\frac{1}{m} \sum_{i=1}^m \sum_{j=n-k+1}^n \log \boldsymbol{p}(w^i_j|w^i_{0:j-1, j+1:n} )
   \label{e:11}
\end{aligned}
\end{equation}

where $w^i_{0:j-1, j+1:n} = w^i_0 \cdot\cdot\cdot w^i_{j-1},w^i_{j+1}..w^i_n$, $\boldsymbol{p}$ is the action decoder head for predicting masked tokens. $w^i_j \in \{0 \cdot\cdot\cdot v-1\}$ is the label for the masked token, $v$ is the required vocabulary size, and $m$ is the number of dialog sessions. Besides, $n$ and $k$ are the length of the input and masked action sequences, respectively. 

\section{Experiments and Results}
We first conduct the simulator evaluation to assess the DaTrans' performance of learning efficiency, the robustness of fine-tuning Corpus, and domain adaptation. Besides, the case study and human evaluation are conducted and the results are presented in Section \ref{human} \& \ref{case} in Appendix. In our experiment, NLU and NLG modules are ignored since the interactions are made with dialog actions. Notably, DaTrans can be equipped with any NLU and NLG models. Two datasets, MultiWoz \cite{budzianowski2018multiwoz} and Schema-Guided dialog (SGD) \cite{rastogi2019towards} are involved. We leverage a public available agenda-based user simulator \cite{zhu2020convlab} setup on MultiWoz. The details of the dataset, implementation, and the user simulator are illustrated in the Appendix.

\subsection{Baseline Agents}
\label{sec:baseline}
We compare the performance of the proposed DaTrans with the state-of-art model JOIE \cite{wang2021collaborative}, vanilla BERT, and its variants of different optimization and fine-tuning settings. \footnote{“optimization” refers to the interactive training process with Reinforcement Learning. “pre-train” means the
process of PLMs trained with massive plain text. Besides, we use both “pre-train” and “fine-tuning” to refer to the self-supervised training
process of BERT with annotated historical action sequences.} 
\textbf{DQN} agent is trained with a deep Q-Network. \textbf{BERT} agent is equipped with BERT as the encoder that replaces the fully connected layer in DQN. \textbf{BERT$_{\mathtt{MWoz}}$} agent is with BERT pre-trained with MLM and NSP tasks on MultiWoz. \textbf{JOIE} agent \cite{wang2021collaborative} is a collaborative multi-agent framework factoring the joint action space and learning each part by a different agent. \textbf{DaTrans$_{\mathtt{MWoz}}$} is our proposed agent that is pre-trained with MLA task as described in Section 3.1 on MultiWoz dataset.

\begin{table}[htbp]
\small
\centering
\caption{\label{tab:simulate-result} The simulation performance of different agents. Succ. denotes the final success rate, Turn and Reward are the average turn and the average reward of the whole training process, respectively.}

\centering
\setlength{\tabcolsep}{2.5mm}{
\begin{tabular}{l ccc} 
\toprule 
Model &Succ.$\uparrow$  &Turn$\downarrow$ &Reward$\uparrow$\\
\midrule 

DaTrans$_{\mathtt{MWoz}}$       
          &\bf{0.84}   &\bf{10.21}  &\bf{27.35}  \\
BERT$_{\mathtt{MWoz}}$      &0.72   &12.14  &14.21   \\
BERT      &0.64   &14.75  &-15.47 \\
DQN       &0.01   &19.51  &-53.66 \\
JOIE-3      &0.38   &15.98  &-21.42 \\
\bottomrule 
\end{tabular}
}
\end{table}

\subsection{Simulator Evaluation}

\begin{figure*}[ht]
\centering  
\subfigure[Main results.]{\label{fmain}\includegraphics[width=0.66\columnwidth]{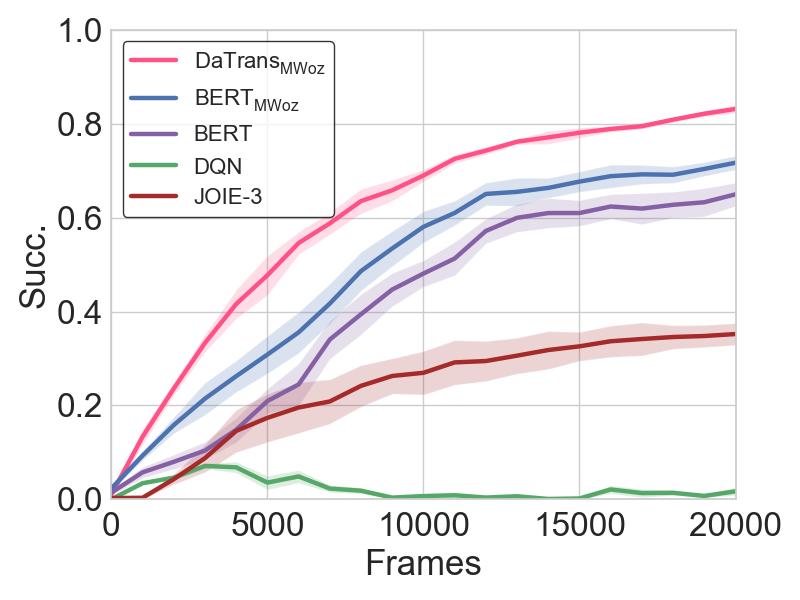}}
\subfigure[Domain adaptation.]{\label{fadapt}\includegraphics[width=0.66\columnwidth]{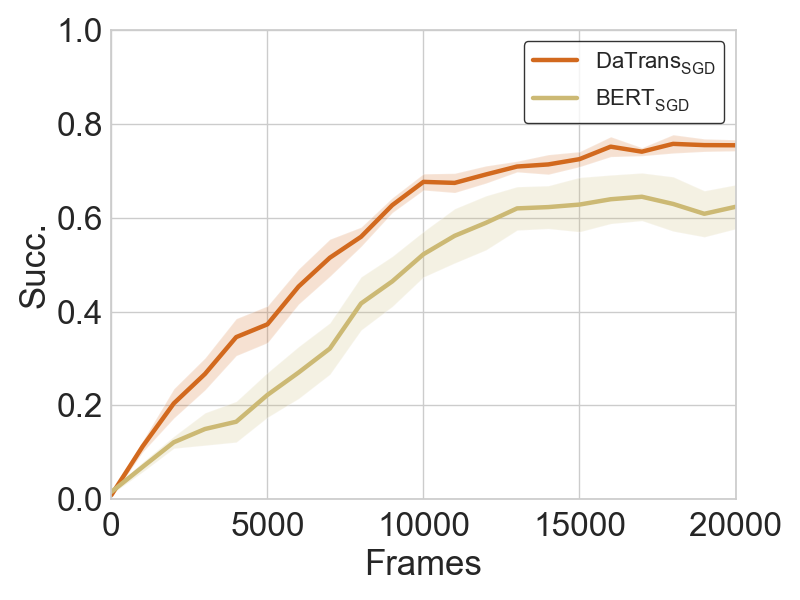}}
\subfigure[Altering fine-tuning corpus.]{\label{fcorpus}\includegraphics[width=0.66\columnwidth]{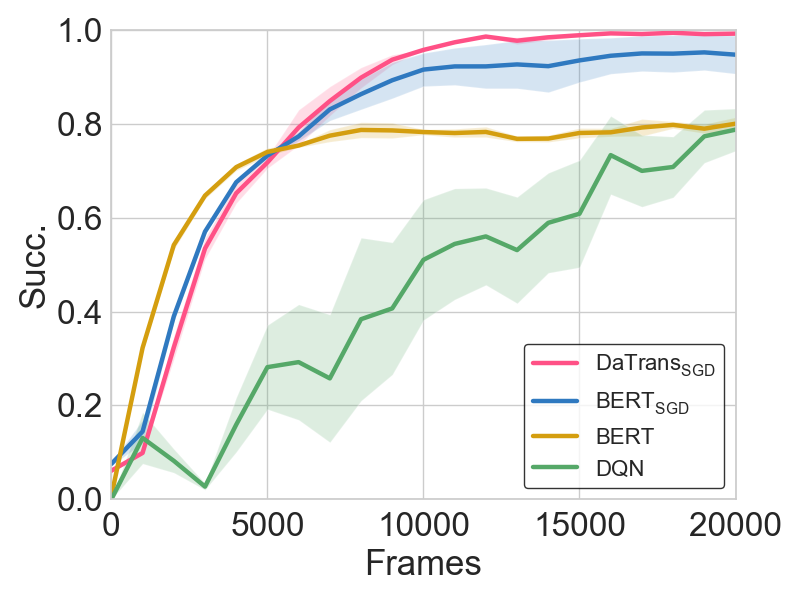}}
\caption{Comparison of the success rate evolving during the training process.}
\end{figure*}

All agents are evaluated with the success rate (Succ.) at the end of the training, average turn (Turn), average reward (Reward). The main simulation results are shown in Table. \ref{tab:simulate-result} and Figure \ref{fmain}. The results indicate that the proposed DaTrans$_{\mathtt{MWoz}}$ learns faster and achieves a better convergence in in-domain evaluation. DaTrans$_{\mathtt{MWoz}}$, pre-trained with the mask last act task (MLA) on the MultiWoz corpus achieves the best Succ. (on average 0.84) with the highest learning efficiency in BERT-based models. The performance of DaTrans$_{\mathtt{MWoz}}$ reveals that our MLA pre-training task can not only encode the characteristics of dialog policy for efficiency improvement but also show better transfer abilities because dropping it BERT$_{\mathtt{MWoz}}$ degrades the performance of  DaTrans$_{\mathtt{MWoz}}$. Additionally, BERT is consistently the worst in BERT-based models, which is not surprising since it is only initialized with official BERT's pre-trained weights without in-domain fine-tuning. The generality of fine-tuning corpus and task, domain awareness, and knowledge transferability of BERT are poor. Furthermore, without any fine-tuning, JOIE and DQN are worse than BERT-based agents. Finally, the comparison results of Turn and Reward are illustrated in Table. \ref{tab:simulate-result}. It depicts that DaTrans$_{\mathtt{MWoz}}$ achieves the shortest average turn and highest average reward, which is consistent with the learning curves in Figure \ref{fmain}.


\noindent\textbf{Effect of fine-tuning Corpus.} We further test the effect of different fine-tuning corpus on the performance. The models are pre-trained on SGD and optimized on MultiWoz to investigate the influence of fine-tuning corpus. We denote DaTrans$_{\mathtt{SGD}}$ as a variant of DaTrans which is pre-trained on SGD and optimized on MultiWoz. We only compared the results of fine-tuning on SGD, because the agents who have fine-tuned on MultiWoz have seen the
dialogue logic of MultiWoz, so it is of little significance to optimize the comparison on MultiWoz. Besides, we don’t optimize the models with RL on SGD because we didn’t find an open-source simulator for SGD. Thus, we only take SGD to explore the effect of corpus and domain adaptation. The core conclusion indicated from Figure \ref{fcorpus} is that DaTrans is robust to the different fine-tuning corpus. Firstly, the proposed MLA pre-training task does better in extracting the knowledge of dialog action sequence, especially the structure information that is invariant over domains. As a consequence, DaTrans$_{\mathtt{SGD}}$ outperforming BERT$_{\mathtt{SGD}}$.

\noindent\textbf{Domain Adaptation.}
To assess the ability for new task adaptation, we compare the agents that continually learn a new domain Restaurant, starting from being well-trained on the other six domains (i.e. Train, Hotel, Hospital, Taxi, Police, Attraction). Figure \ref{fadapt} shows the performances of new task adaptation for dialog policy learning. The results confirm that DaTrans pre-trained with masked last action task is capable of quickly adapting to the new environment compared to DaTrans$_{\mathtt{SGD}}$ and BERT$_{\mathtt{SGD}}$. Besides, pre-training counts because removing it (BERT) damages the results. 

\section{Conclusion and Future Work}
In this paper, we investigate the pre-trained language model enhancing the reinforcement learning agent for dialog policy learning. We propose DaTrans, which is equipped with a new fine-tuning task that masks the last dialog action to extract the dialog logic for efficient dialog policy learning. The evaluation results show the effectiveness of the proposed DaTrans in terms of learning efficiency and domain adaptation ability. 

\section*{Limitations}

Due to the high cost of interactions with human users, the dialog policy model was trained in a simulated environment rather than real-world scenarios. Our approach is able to construct a highly responsive dialog system because it shortens the required interaction turns, and reduces labour costs associated with interactive training with human users. However, it is worth noting that the model optimized in our experiments may not be suitable for dealing with real-world users, thus simulation evaluation results alone are not sufficient to prove DaTrans's superiority. Despite this limitation, as there are few studies dedicated to investigating PLMs advanced dialog policy learning, We hope that DaTrans will inspire further research in this field in the future.
\section*{Acknowledgements}
We appreciate the constructive and insightful comments provided by the anonymous reviewers. This research work is partially supported by CUHK under Project No. 3230377.

\bibliography{anthology,custom}

\begin{thebibliography}{20}
\expandafter\ifx\csname natexlab\endcsname\relax\def\natexlab#1{#1}\fi

\bibitem[{Brys et~al.(2015)Brys, Harutyunyan, Suay, Chernova, Taylor, and
  Now{\'e}}]{brys2015reinforcement}
Tim Brys, Anna Harutyunyan, Halit~Bener Suay, Sonia Chernova, Matthew~E Taylor,
  and Ann Now{\'e}. 2015.
\newblock Reinforcement learning from demonstration through shaping.
\newblock In \emph{Twenty-Fourth International Joint Conference on Artificial
  Intelligence}.

\bibitem[{Budzianowski et~al.(2018)Budzianowski, Wen, Tseng, Casanueva, Ultes,
  Ramadan, and Ga{\v{s}}i{\'c}}]{budzianowski2018multiwoz}
Pawe{\l} Budzianowski, Tsung-Hsien Wen, Bo-Hsiang Tseng, I{\~n}igo Casanueva,
  Stefan Ultes, Osman Ramadan, and Milica Ga{\v{s}}i{\'c}. 2018.
\newblock \href {https://doi.org/10.18653/v1/D18-1547} {{M}ulti{WOZ} - a
  large-scale multi-domain {W}izard-of-{O}z dataset for task-oriented dialogue
  modelling}.
\newblock In \emph{Proceedings of the 2018 Conference on Empirical Methods in
  Natural Language Processing}, pages 5016--5026, Brussels, Belgium.
  Association for Computational Linguistics.

\bibitem[{Cederborg et~al.(2015)Cederborg, Grover, Isbell, and
  Thomaz}]{cederborg2015policy}
Thomas Cederborg, Ishaan Grover, Charles~L Isbell, and Andrea~L Thomaz. 2015.
\newblock Policy shaping with human teachers.
\newblock In \emph{Twenty-Fourth International Joint Conference on Artificial
  Intelligence}.

\bibitem[{Devlin et~al.(2019)Devlin, Chang, Lee, and
  Toutanova}]{devlin2018bert}
Jacob Devlin, Ming-Wei Chang, Kenton Lee, and Kristina Toutanova. 2019.
\newblock \href {https://doi.org/10.18653/v1/N19-1423} {{BERT}: Pre-training of
  deep bidirectional transformers for language understanding}.
\newblock In \emph{Proceedings of the 2019 Conference of the North {A}merican
  Chapter of the Association for Computational Linguistics: Human Language
  Technologies, Volume 1 (Long and Short Papers)}, pages 4171--4186,
  Minneapolis, Minnesota. Association for Computational Linguistics.

\bibitem[{Geishauser et~al.(2022)Geishauser, van Niekerk, Lin, Lubis, Heck,
  Feng, and Gasic}]{geishauser2022dynamic}
Christian Geishauser, Carel van Niekerk, Hsien-Chin Lin, Nurul Lubis, Michael
  Heck, Shutong Feng, and Milica Gasic. 2022.
\newblock Dynamic dialogue policy for continual reinforcement learning.
\newblock In \emph{Proceedings of the 29th International Conference on
  Computational Linguistics}, pages 266--284.

\bibitem[{Gulyaev et~al.(2020)Gulyaev, Elistratova, Konovalov, Kuratov,
  Pugachev, and Burtsev}]{gulyaev2020goal}
Pavel Gulyaev, Eugenia Elistratova, Vasily Konovalov, Yuri Kuratov, Leonid
  Pugachev, and Mikhail Burtsev. 2020.
\newblock Goal-oriented multi-task bert-based dialogue state tracker.
\newblock \emph{arXiv preprint arXiv:2002.02450}.

\bibitem[{Jhunjhunwala et~al.(2020)Jhunjhunwala, Bryant, and
  Shah}]{jhunjhunwala2020multi}
Megha Jhunjhunwala, Caleb Bryant, and Pararth Shah. 2020.
\newblock Multi-action dialog policy learning with interactive human teaching.
\newblock In \emph{Proceedings of the 21th Annual Meeting of the Special
  Interest Group on Discourse and Dialogue}, pages 290--296.

\bibitem[{Kwan et~al.(2023)Kwan, Wang, Wang, and Wong}]{kwan_survey_2023}
Wai-Chung Kwan, Hong-Ru Wang, Hui-Min Wang, and Kam-Fai Wong. 2023.
\newblock A survey on recent advances and challenges in reinforcement learning
  methods for task-oriented dialogue policy learning.
\newblock \emph{Machine Intelligence Research}, 20(3):318--334.

\bibitem[{Lee et~al.(2019)Lee, Zhu, Takanobu, Zhang, Zhang, Li, Li, Peng, Li,
  Huang, and Gao}]{lee2019convlab}
Sungjin Lee, Qi~Zhu, Ryuichi Takanobu, Zheng Zhang, Yaoqin Zhang, Xiang Li,
  Jinchao Li, Baolin Peng, Xiujun Li, Minlie Huang, and Jianfeng Gao. 2019.
\newblock \href {https://doi.org/10.18653/v1/P19-3011} {{C}onv{L}ab:
  Multi-domain end-to-end dialog system platform}.
\newblock In \emph{Proceedings of the 57th Annual Meeting of the Association
  for Computational Linguistics: System Demonstrations}, pages 64--69,
  Florence, Italy. Association for Computational Linguistics.

\bibitem[{Li et~al.(2020)Li, Kiseleva, and de~Rijke}]{li2020rethinking}
Ziming Li, Julia Kiseleva, and Maarten de~Rijke. 2020.
\newblock \href {https://doi.org/10.18653/v1/2020.findings-emnlp.316}
  {Rethinking supervised learning and reinforcement learning in task-oriented
  dialogue systems}.
\newblock In \emph{Findings of the Association for Computational Linguistics:
  EMNLP 2020}, pages 3537--3546, Online. Association for Computational
  Linguistics.

\bibitem[{Peng et~al.(2017)Peng, Li, Li, Gao, Celikyilmaz, Lee, and
  Wong}]{peng2017composite}
Baolin Peng, Xiujun Li, Lihong Li, Jianfeng Gao, Asli Celikyilmaz, Sungjin Lee,
  and Kam-Fai Wong. 2017.
\newblock \href {https://doi.org/10.18653/v1/D17-1237} {Composite
  task-completion dialogue policy learning via hierarchical deep reinforcement
  learning}.
\newblock In \emph{Proceedings of the 2017 Conference on Empirical Methods in
  Natural Language Processing}, pages 2231--2240, Copenhagen, Denmark.
  Association for Computational Linguistics.

\bibitem[{Peng et~al.(2020)Peng, Zhu, Li, Li, Li, Zeng, and Gao}]{peng2020few}
Baolin Peng, Chenguang Zhu, Chunyuan Li, Xiujun Li, Jinchao Li, Michael Zeng,
  and Jianfeng Gao. 2020.
\newblock \href {https://doi.org/10.18653/v1/2020.findings-emnlp.17} {Few-shot
  natural language generation for task-oriented dialog}.
\newblock In \emph{Findings of the Association for Computational Linguistics:
  EMNLP 2020}, pages 172--182, Online. Association for Computational
  Linguistics.

\bibitem[{Rastogi et~al.(2020)Rastogi, Zang, Sunkara, Gupta, and
  Khaitan}]{rastogi2019towards}
Abhinav Rastogi, Xiaoxue Zang, Srinivas Sunkara, Raghav Gupta, and Pranav
  Khaitan. 2020.
\newblock \href {https://doi.org/10.1609/aaai.v34i05.6394} {Towards scalable
  multi-domain conversational agents: The schema-guided dialogue dataset}.
\newblock In \emph{Proceedings of the {AAAI} Conference on Artificial
  Intelligence}, volume~34, pages 8689--8696.
\newblock Number: 05.

\bibitem[{Su et~al.(2016)Su, Gasic, Mrksic, Rojas-Barahona, Ultes, Vandyke,
  Wen, and Young}]{su2016continuously}
Pei-Hao Su, Milica Gasic, Nikola Mrksic, Lina Rojas-Barahona, Stefan Ultes,
  David Vandyke, Tsung-Hsien Wen, and Steve Young. 2016.
\newblock Continuously learning neural dialogue management.
\newblock \emph{arXiv preprint arXiv:1606.02689}.

\bibitem[{Wang et~al.(2020)Wang, Peng, and Wong}]{wang2020learning}
Huimin Wang, Baolin Peng, and Kam-Fai Wong. 2020.
\newblock Learning efficient dialogue policy from demonstrations through
  shaping.
\newblock In \emph{Proceedings of the 58th Annual Meeting of the Association
  for Computational Linguistics}, pages 6355--6365.

\bibitem[{Wang and Wong(2021)}]{wang2021collaborative}
Huimin Wang and Kam-Fai Wong. 2021.
\newblock A collaborative multi-agent reinforcement learning framework for
  dialog action decomposition.
\newblock In \emph{Proceedings of the 2021 Conference on Empirical Methods in
  Natural Language Processing}, pages 7882--7889.

\bibitem[{Wolf et~al.(2020)Wolf, Chaumond, Debut, Sanh, Delangue, Moi, Cistac,
  Funtowicz, Davison, Shleifer et~al.}]{wolf2020transformers}
Thomas Wolf, Julien Chaumond, Lysandre Debut, Victor Sanh, Clement Delangue,
  Anthony Moi, Pierric Cistac, Morgan Funtowicz, Joe Davison, Sam Shleifer,
  et~al. 2020.
\newblock Transformers: State-of-the-art natural language processing.
\newblock In \emph{Proceedings of the 2020 Conference on Empirical Methods in
  Natural Language Processing: System Demonstrations}, pages 38--45.

\bibitem[{Yang et~al.(2021)Yang, Li, and Quan}]{yang2020ubar}
Yunyi Yang, Yunhao Li, and Xiaojun Quan. 2021.
\newblock Ubar: Towards fully end-to-end task-oriented dialog system with
  gpt-2.
\newblock In \emph{Proceedings of the AAAI Conference on Artificial
  Intelligence}, volume~35, pages 14230--14238.

\bibitem[{Young et~al.(2013)Young, Ga{\v{s}}i{\'c}, Thomson, and
  Williams}]{young2013pomdp}
Steve Young, Milica Ga{\v{s}}i{\'c}, Blaise Thomson, and Jason~D Williams.
  2013.
\newblock Pomdp-based statistical spoken dialog systems: A review.
\newblock \emph{Proceedings of the IEEE}, 101(5):1160--1179.

\bibitem[{Zhu et~al.(2020)Zhu, Zhang, Fang, Li, Takanobu, Li, Peng, Gao, Zhu,
  and Huang}]{zhu2020convlab}
Qi~Zhu, Zheng Zhang, Yan Fang, Xiang Li, Ryuichi Takanobu, Jinchao Li, Baolin
  Peng, Jianfeng Gao, Xiaoyan Zhu, and Minlie Huang. 2020.
\newblock \href {https://doi.org/10.18653/v1/2020.acl-demos.19} {{C}onv{L}ab-2:
  An open-source toolkit for building, evaluating, and diagnosing dialogue
  systems}.
\newblock In \emph{Proceedings of the 58th Annual Meeting of the Association
  for Computational Linguistics: System Demonstrations}, pages 142--149,
  Online. Association for Computational Linguistics.

\end{thebibliography}
\bibliographystyle{acl_natbib}



\clearpage
\appendix
\section{Dataset}
Two datasets are involved: 1) MultiWoz \cite{budzianowski2018multiwoz}, a large-scale fully annotated corpus of human-human conversations; 2) Schema-Guided dialog (SGD) \cite{rastogi2019towards}, multi-domain, task-oriented conversations between a human and a virtual assistant. MultiWOz contains 8,434 pieces of corpus covering 9 domains, while SGD consists of 16,142 pieces of dialog sessions involving 16 domains. 

\section{Implementation Details.}
\label{Implementation_Details.} 
We adopt BERT$_{base}$ (uncased) with default hyperparameters in Huggingface Transformers \cite{wolf2020transformers} as the backbone transformer encoder model. We pre-train and optimize BERT-based models on one RTX 2080Ti GPU and GTX TITAN X. The pre-training batch size is 8. The learning rate for the BERT-based model is 0.00003. The action decoder of DaTrans is a linear layer with 400 output units corresponding to the 400 action candidates. Meanwhile, we set the discount factor $\gamma$ as 0.9. Besides, we apply the rule-based agent from ConvLab \cite{lee2019convlab} to warm start the policy with 1000 dialog epochs. 

\section{User Simulator}
\label{Simulator}
We leverage a public available agenda-based user simulator \cite{zhu2020convlab} setup on MultiWoz. During training, the simulator initializes with a user goal and takes a system action as input and outputs the user action with a reward. The reward is set as -1 for each turn to encourage short turns and a positive reward ($2\cdot T$) for successful dialog or a negative reward of $-T$ for failed one, where $T$ (set as 40) is the maximum number of turns in each dialog. A dialog is considered successful only if the agent helps the user simulator accomplish the goal and satisfies all the user's search constraints. 

\section{Human Evaluation}
\label{human}
\begin{table}[htbp] 
\centering
\small
\caption{\label{tab:human-evaluation} The Human performance of different agents. The evaluation is conducted at 10000 epochs in Figure \ref{fmain} for all agents. Succ. denotes success rate.}
\centering
\setlength{\tabcolsep}{8.0mm}{
\begin{tabular}{l c}
\toprule 
Model&Succ.$\uparrow$ \\
\midrule 
DaTrans$_{\mathtt{MWoz}}$   &0.68 \\
BERT$_{\mathtt{MWoz}}$      &0.58  \\
BERT                        &0.46 \\
DQN                         &0.00 \\
JOIE-3                      &0.24 \\
\bottomrule
\end{tabular}}
\end{table}
We further conduct a human evaluation to validate the simulation results. We choose the agents trained with 10000 epochs. Before the test, all evaluators are instructed to interact with the agents to achieve their goals. In each session, a randomly selected goal and a random agent are assigned to a user. They can terminate the dialog if they think the session is doomed to fail. At the end of each session, the user is required to judge if the dialog is a success or a failure. We collect 50 conversations for each agent. The results are illustrated in Table. \ref{tab:human-evaluation}. We see that the human evaluation results further convince the simulator evaluation.

\section{Case Study}
\label{case}
\begin{figure}[t]
\centering
\includegraphics[width=0.8\columnwidth]{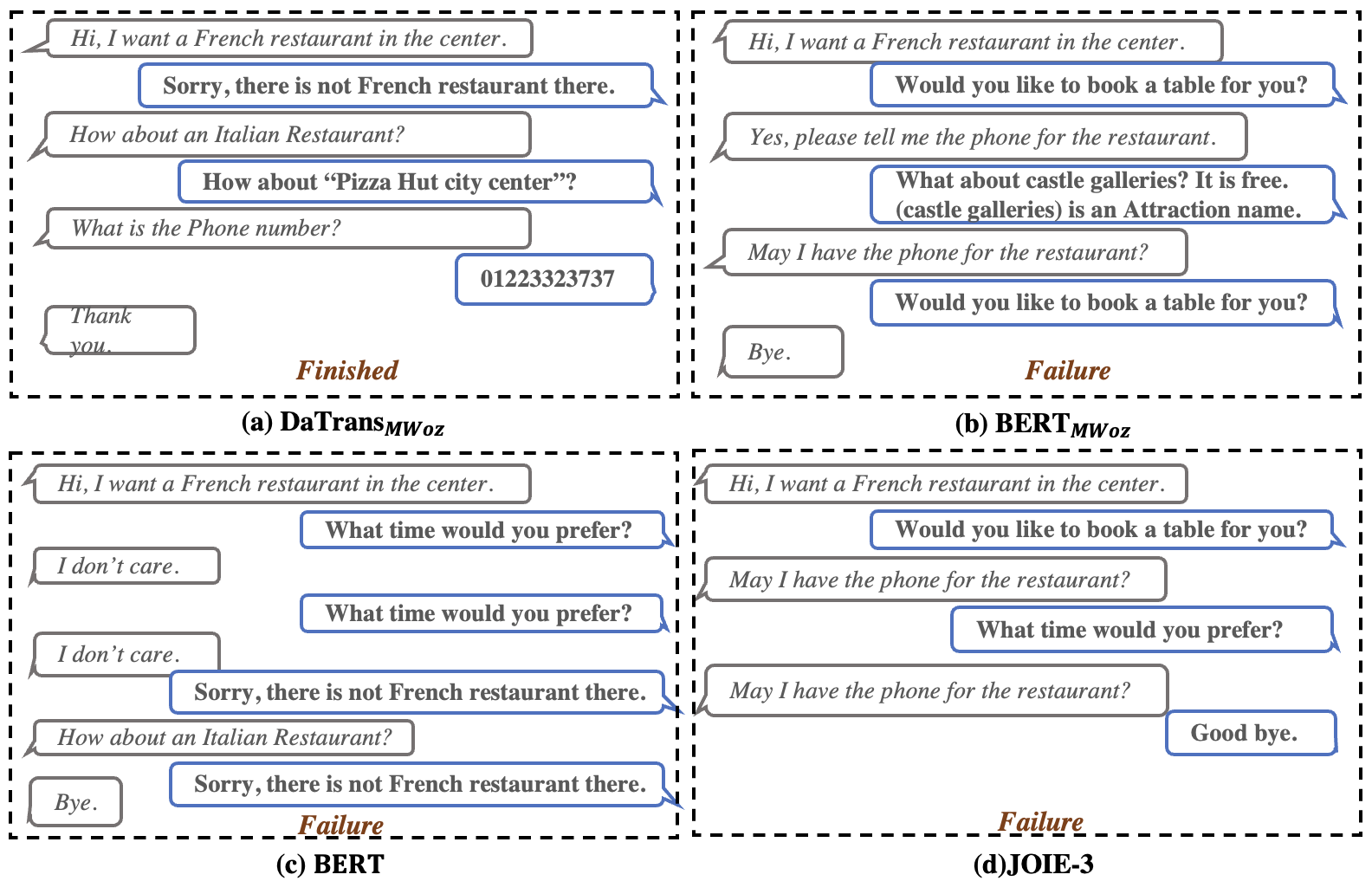}
\caption{Sampled dialogue examples generated by DaTrans$_{\mathtt{MWoz}}$, BERT$_{\mathtt{MWoz}}$, BERT, DQN, JOIE\-3. The grey boxes convey the queries from the users while the blue boxes are the responses from the agents. At the bottom of the boxes, we marked whether the session is successful or not.}
\label{fexa}
\end{figure}

To further explore the performance of the agents after training, we randomly sampled some real examples generated for a shared restaurant goal. From the samples placed in Fig. \ref{fexa}, some explicable clues are found. In this example, BERT$_{\mathtt{MWoz}}$ fails because it makes mistakes in the restaurant's dialogue logic though it recognizes the right domain. Besides, the response involving “castle galleries” indicates BERT$_{\mathtt{MWoz}}$ suffers from disturbance from other task Attraction. As for BERT and JOIE\-3, it seems that the knowledge regarding restaurant has not been mastered. Only DaTrans$_{\mathtt{MWoz}}$ systematically handles the issues by taking reasonable actions.

\end{document}